\documentclass[review]{cvpr}
\usepackage{cuted}
\usepackage{capt-of}
\usepackage{times}
\usepackage{epsfig}
\usepackage{graphicx}
\usepackage{amsmath}
\usepackage{amssymb}
\usepackage{booktabs}
\usepackage{bm}
\usepackage{algorithm}  
\usepackage{algorithmic}
\usepackage{amsmath}  
\usepackage{bbding}

\usepackage[pagebackref=true,breaklinks=true,colorlinks,bookmarks=false]{hyperref}

\begin{document}

\title{Inter-layer Transition in Neural Architecture Search}
\author{%
  Benteng Ma\textsuperscript{1,2} 
  \and
  Jing Zhang\textsuperscript{2}
  \and
  Yong Xia\textsuperscript{1}
  \and
 Dacheng Tao\textsuperscript{2}
  \and 
  \textsuperscript{1}Northwestern Polytechnical University, China \\
  \textsuperscript{2}The University of Sydney, Australia \\
  
}

\twocolumn[{
\maketitle
\begin{figure}[H]
    \hsize=\textwidth
    \centering
    \includegraphics[width=2.0\linewidth]{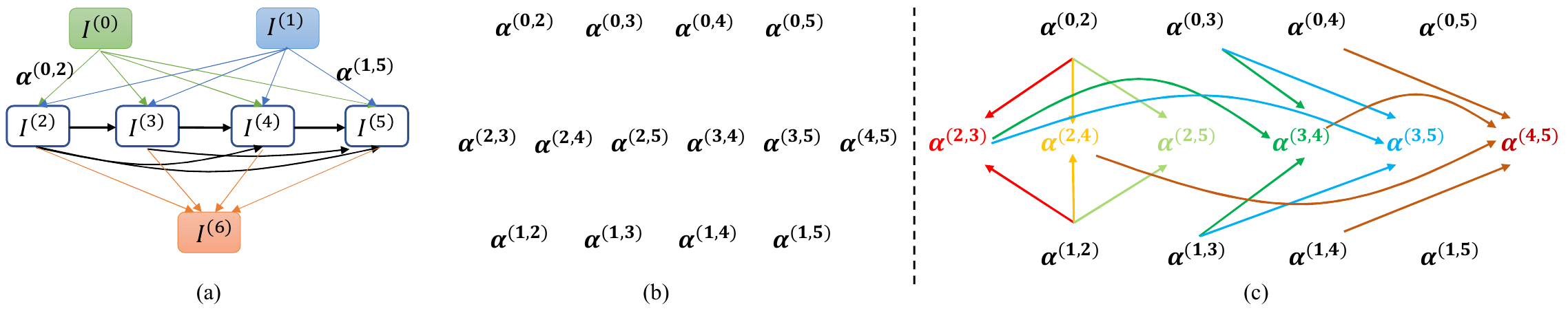}
    \caption{(a) An exemplar directed acyclic graph of a cell. The blue and green nodes are outputs of previous cells. The orange node is the output of the current cell. The white nodes are features learned by layers in the current cell (\ie, the directed edges), where $\bm{\alpha^{(i,j)}}$ represents the categorical probability of operations on edge $(i,j)$. (b) Existing methods optimize $\bm{\alpha^{(i,j)}}$ independently. (c) Our method learns the inter-layer transition to model the dependency between topologically connected edges in  an attentive probability transition manner.}
    \label{fig:opening}
\end{figure}
}]
\maketitle

\begin{abstract}
Differential Neural Architecture Search (NAS) methods represent the network architecture as a repetitive proxy directed acyclic graph (DAG) and optimize the network weights and architecture weights alternatively in a differential manner. However, existing methods model the architecture weights on each edge (i.e., a layer in the network) as statistically independent variables, ignoring the dependency between edges in DAG induced by their directed topological connections. In this paper, we make the first attempt to investigate such dependency by proposing a novel Inter-layer Transition NAS method. It casts the architecture optimization into a sequential decision process where the dependency between the architecture weights of connected edges is explicitly modeled. Specifically, edges are divided into inner and outer groups according to whether or not their predecessor edges are in the same cell. While the architecture weights of outer edges are optimized independently, those of inner edges are derived sequentially based on the architecture weights of their predecessor edges and the learnable transition matrices in an attentive probability transition manner. Experiments on five benchmarks confirm the value of modeling inter-layer dependency and demonstrate the proposed method outperforms state-of-the-art methods. Code will be  available at \url{https://github.com/btma48/ITNAS}.
\end{abstract}
\section{Introduction}
Recently, the progress in computer vision research has been advanced by deep learning dramatically. 
The essential ingredient of deep learning is the design and optimization of the target deep neural networks. While the early stages of deep learning placed great emphasis on feature engineering \cite{he2016deep,hu2018squeeze,xie2017aggregated}, the focus has now migrated to architecture surgery. However, manually designing neural network architectures entails both specialized expertise and tremendous computational resources. Neural Architecture Search (NAS) presents a promising alternative approach to alleviating this affliction by automatically searching architectures superior to those by manual design \cite{elsken2019neural}. NAS has achieved remarkable success in image recognition\cite{fang2020densely}, semantic segmentation \cite{liu2019auto,nekrasov2019fast}, and object detection \cite{xu2019auto,guo2020hit,tan2020efficientdet}.

The linchpin to NAS methodology is to build a massive network architecture space, devise an efficient algorithm to explore this space, and uncover the optimal architecture with a portfolio of training data and constraints \cite{elsken2019neural}. Ordinarily, the cell to be searched is defined as a directed acyclic graph, as illustrated in Figure~\ref{fig:opening}(a). And NAS methods formulate the search problem as a bi-level optimization problem, \ie, model training as the lower-level optimization problem and architectural search as the upper-level one \cite{liu2018darts}. Such an optimization problem has been shown to be computationally difficult in previous studies due to the fact that the bi-level optimization solutions dynamically interact with and affect each other. Extensive research has been conducted to address the bi-level optimization problem, \eg, reinforcement learning (RL)-based methods \cite{zoph2016neural,zoph2018learning,baker2016designing,zhong2018practical}, evolutionary algorithms (EA) \cite{real2019regularized,real2017large}, and gradient-based methods \cite{liu2018darts,he2020milenas,luo2018neural,xie2018snas}.

In both RL-based and EA-based methodologies, the validation precision of a multitude of network architecture candidates is expected in their search strategies, which is computationally expensive and extremely time-consuming. For example, RL-based methods are generally established as a Markov decision process and have often used validation precision as a reward to optimize architecture generators in each iteration. EA-based methods decide whether to select or remove an architecture candidate from the population until the architecture is chosen and evaluated. In these methods, thousands of GPU days are devoted to finding the optimal architecture on the CIFAR-10 dataset, which is inefficient and overwhelming. Recent works, epitomized by DARTS \cite{liu2018darts}, have erected a new lineage of methodologies which relax the discrete search space to be continuous and differentiable. These methods employ gradient-based strategies to optimize neural architectures, resulting in orders of magnitude computations reductions in search costs, \ie, from thousands of GPU days to a few GPU hours. It opens a computationally feasible route for architecture search. In DARTS \cite{liu2018darts}, the entire computation graph composed of repeating computation cells is learned together. At the end of the search phase, a pruning process is applied to the edge connections and their associated operations within the cells by keeping those with the highest architecture weights, which represent the strength of edge connections. 

While DARTS achieves impressive performance, these methods model the architecture weights on each edge as statistically independent variables and ignore the dependency among edges in DAG induced by their directed topological connections, as shown in Figure~\ref{fig:opening}(b). In addition, the hard pruning process after search is heuristic and sub-optimal. To address these issues, we make the first attempt to investigate the dependency by proposing a novel Inter-layer Transition NAS (ITNAS) method. It adopts an architecture transition learning paradigm that can explicitly model the dependency between the architecture weights of predecessor and posterior edges. As shown in Figure~\ref{fig:opening}(c), we divide the edges in each cell into two groups, \ie, inner (denoted in color) and outer group, according to whether or not their predecessor edges are in the same cell. While the architecture weights of outer edges are optimized independently, those of inner edges are derived sequentially based on the weights of their predecessor edges and associated learnable transition matrices in an attentive probability transition manner. Moreover, a novel transition-induced iterative edge pruning method is proposed to prune edges and generate the final architecture.

The contribution of this paper can be summarized as follows. First, we propose a novel NAS method to model the inter-layer dependency for the first time. With learnable transition matrices and attention vectors, the weights of inner edges can be derived from the weights of their predecessor edges sequentially rather than being learned independently. Second, we propose a new transition-induced iterative edge pruning strategy that can progressively prune edges in the context of the architecture transition paradigm. Third, extensive experiments on five benchmarks demonstrate that the proposed ITNAS can search better architectures than state-of-the-art methods, validating that modeling the inter-layer dependency matters in NAS.

\section{Related Work}
Recently, the automatic design of neural architectures,  a.k.a. Neural Architecture Search (NAS), has attracted increasing attention. NAS methods are commonly classified from three perspectives, \ie, the search space, search strategy, and performance estimation strategy \cite{elsken2019neural}. The majority of earlier NAS works typically choose the macro search space, where the entire network is discovered \cite{zoph2016neural}.  For example, the pioneering NAS method relies on the evolutionary algorithm, which simultaneously optimizes architectures and network weights using vast computational resources  \cite{real2019regularized,real2017large}. Later, reinforcement learning-based approaches have been proposed for NAS \cite{zoph2016neural,zoph2018learning,baker2016designing,zhong2018practical}. They formulate the design process of a neural network as a sequence of actions and regard the model accuracy as a reward. These methods usually use a recurrent network as a controller \cite{zoph2018learning} to generate the model representation of a sampled neural network designed for a given task.

In order to search in continuous domains, DARTS \cite{liu2018darts} proposes a continuous relaxation of the architecture representation then leverages the efficient gradient back-propagation to optimize the architecture weights and network weights alternatively. As a result, DARTS is able to find good convolutional architectures at a much less computational cost, making NAS broadly accessible. Owed to the success of DARTS, several extensions have been proposed. SNAS \cite{xie2018snas} optimizes the parameters of a joint distribution for the search space in a cell. The authors propose a search gradient method that optimizes the same objective as RL-based NAS but leads to more efficient structural decisions. P-DARTS \cite{chen2019progressive} attempts to overcome the depth gap issue between the search phase and evaluation phase, which is accomplished by increasing the depth of searched architectures gradually during the training phase. PC-DARTS \cite{xu2019pc} leverages the redundancy in the network space and only samples a subset of channels in super-net during search to reduce computation cost. MileNAS \cite{he2020milenas} incorporates both bi-level and single-level optimization schemes into a mixed-level framework. SGAS \cite{li2020sgas} chooses and prunes candidate operations in a greedy fashion by dividing the search procedure into subproblems. All these differential NAS methods model the architecture weights of edges in a cell as independent variables and optimize them simultaneously under a framework. However, some edges (\ie, inner edges) within the cell are actually not independent from others due to the directed topological connections between them and their predecessor edges. In contrast to the methods mentioned above, we make the first attempt to investigate the dependency between edges in DAG and propose a novel NAS method by modeling the inter-layer transition. 

Most differential NAS methods \cite{liu2018darts,xu2019pc} adopt a hard pruning strategy to derive the final architecture from the searched network. First, only the operation with the largest architecture weight for each edge is retained. Then, for each intermediate node, only the two edges with the largest edge importance are retained. Recently, ASAP \cite{noy2020asap} adopts a progressive pruning strategy to prune the operations whose architecture weights are below a threshold that is updated gradually. However, these heuristic pruning strategies prune all weak edges without considering the dependency between them and their descendent edges induced by the directed topological connections, resulting in a sub-optimal solution. By contrast, we propose a new transition-induced iterative edge pruning strategy method that can progressively prune edges while updating the edge importance of their descendent edges dynamically by leveraging the inter-layer transition matrices.

\section{Method}
\subsection{Preliminary}
Architecture search in DARTS focuses on searching optimal repeating patterns of structure in the network, which are called cells. This formulation is inspired by the observation that the advanced manually designed deep neural networks always consist of a few computational blocks that are stacked sequentially to form the final structure of the network. No matter how many blocks the network contains, the structures within each block are repetitive. Thus, to learn one or two structures, which are represented as unit cells, is sufficient to design an entire network. 

Given an architecture space, we can represent the architecture of a cell as a directed acyclic graph (DAG), \ie, $\bm{\alpha=(I, E)}$, where $\mathbf{I}=\{\mathbf{I}^{(i)}\}$ is a set of nodes that represents the feature maps in the neural network and $E$ is an edge set. The directed edge $(i,j)\in E$ denotes a layer that transforms the feature map from the node $\mathbf{I}^{(i)}$ to $\mathbf{I}^{(j)}$. Since the DAG is composed of an ordered sequence of nodes, where each edge can only point from a low-indexed node to a high-indexed node, thereby we have $i<j$ for each edge. The output of the intermediate node is computed as:
\begin{equation}
    \mathbf{I}^{(j)} = \sum_{i<j} \hat{o}^{(i,j)}(\mathbf{I}^{(i)}),
    \label{eq:tensorcal_o}
\end{equation}
where $\hat{o}^{(i,j)}\in \mathcal{O}$ is the selected operation (\eg, convolution or pooling) at edge $(i,j)$ from a operation set $\mathcal{O}$ containing $K$ operations. An architecture weight vector $\bm{\alpha^{(i,j)}}$ is introduced to represent the probability distribution of the operation categorical choice on the edge $(i,j)$. Thus, the differential NAS methods target to optimize the $\bm{\alpha^{(i,j)}}$ to search the optimal operation $\hat{o}^{(i,j)}$. In DARTS \cite{liu2018darts} and its variants \cite{xu2019pc}, the optimal architecture is searched from the discrete search space by relaxing the selection of the operations to a continuous problem. By sampling all the possible operations $o$ for edge $(i,j)$, Eq.~\eqref{eq:tensorcal_o} can be re-written as:
\begin{equation}
    \mathbf{I}^{(j)}=\sum_{i<j} \left( \sum_{o\in\mathcal{O}}\frac{\exp ({\alpha_{o}^{(i,j)}})}{\sum_{o'\in \mathcal{O}} \exp ({\alpha_{o'}^{(i,j)}})}\cdot o(\mathbf{I}^{(i)}) \right).
    \label{eq:tensorcal_o_prob}
\end{equation}
Note that we apply the softmax to all architecture weights $\alpha_{o}^{(i,j)}$ of operation $o$ to obtain the probability distribution. The output of a node is the weighted sum of the outputs from all candidate operations. Then, searching the optimal architectural parameters $\bm{\alpha}$ can be formulated as a bi-level optimization problem:
\begin{align}
    &\mathop{\text{min}}\limits_{\bm{\alpha}} \ \ \    \mathcal{L}_{val}(\bm{\omega^*(\alpha)},\bm{\alpha}) \nonumber \\
    \text{s.t.} \ \ \ \ &\bm{\omega^*(\alpha)} = \mathop{\text{argmin}}\limits_{\bm{\omega}} \mathcal{L}_{train}(\bm{\omega},\bm{\alpha}),
    \label{eq:bilevel_opt}
\end{align}
where $\bm{\omega}$ denotes the learnable weights of the supernet, $\mathcal{L}_{train}$ and $\mathcal{L}_{val}$ denote the training loss and validation loss, respectively. The search is carried out by optimizing the supernet under the continuous operation relaxation. 

\subsection{Anneal Gumbel Softmax}

As shown in Figure \ref{fig:opening}, the first two nodes $\mathbf{I}^{(0)}$ and $\mathbf{I}^{(1)}$ are outputs from previous two cells, the intermediate nodes from $\mathbf{I}^{(2)}$ to $\mathbf{I}^{(5)}$ are calculated according to Eq.~\eqref{eq:tensorcal_o_prob}, and the node $\mathbf{I}^{(6)}= \bigcup_{i=2}^5 \mathbf{I}^{(i)}$ is the output of the current cell, which is the channel-wise concatenation of the four intermediate nodes. The operation $o^{(i,j)}$ of predecessors edge is selected according to the architecture weight $\bm{\alpha^{(i,j)}}$ based on a specific rule. The most straightforward way is argmax that selects the one with the largest architecture weight. After the edge operation selection, the architecture weight vector $\bm{\alpha^{(i,j)}}$ can be represented by an one-hot vector $\hat{\mathbf{Z}}^{(i,j)}$ and Eq.~\eqref{eq:tensorcal_o_prob} can be re-written as:
\begin{align}
    &\mathbf{I}^{(j)} = \sum_{i<j}\sum_{k=0}^{K-1}\hat{\mathbf{Z}}^{(i,j)}_k \cdot o_k(\mathbf{I}^{(i)}), \nonumber \\
    \text{s.t.}\ \ &\hat{\mathbf{Z}}^{(i,j)}= \text{one-hot}\left( \underset{k}{\text{argmax}}\{\alpha_k^{(i,j)}\} \right).
    \label{equation:onehot}
\end{align}
The typical way to represent the probability distribution of a discrete random variable is to employ a categorical distribution. However, it will be extremely challenging to back-propagate errors since the discrete representation $\hat{\mathbf{Z}}^{(i,j)}$ is not differentiable. Thereby, the reparameterization trick is used to relax the discrete architecture distribution by using the concrete distribution \cite{maddison2016concrete}, \ie,
\begin{equation}
    Z^{(i,j)}_k = \frac{\exp\left(\left(\log \alpha^{(i,j)}_k + G^{(i,j)}_k\right)/\tau\right)}{\sum_{k'=0}^{K-1}\exp \left(\left(\log \alpha^{(i,j)}_{k'}+G^{(i,j)}_{k'}\right)/\tau\right)},
\label{eq:intermediate}
\end{equation}
where $Z^{(i,j)}_k$ is the $k$-th element of $\mathbf{Z}^{(i,j)}$, \ie, the softened one-hot vector for operation selection at edge $(i,j)$, $G^{(i,j)}_k = -\log(-\log(U^{(i,j)}_k))$ is the $k$-th Gumbel random variable, $U^{(i,j)}_k$ is a uniform random variable, and $\tau$ is the temperature of the softmax, which is steadily annealed to be closed to zero during the architecture searching phase. In this way, the non-differentiable $\hat{\mathbf{Z}}^{(i,j)}$ in Eq.~\eqref{equation:onehot} can be refactored into a differentiable $\mathbf{Z}^{(i,j)}$ in Eq.~\eqref{eq:intermediate}. And we can get the following equation as been proven in \cite{maddison2016concrete}:
\begin{equation}
    p\left(\mathop{\text{lim}}\limits_{\tau \to 0} Z^{(i,j)}_k=1\right) = \frac{\alpha^{(i,j)}_k}{\sum_{k'=0}^{K-1}\alpha^{(i,j)}_{k'}},
\end{equation}
\ie, the above relaxation is asymptotically unbiased. When $\tau \rightarrow \infty$, each element in $\mathbf{Z}^{(i,j)}$ will be the same. When $\tau \rightarrow 0$, $\mathbf{Z}^{(i,j)}$ will be an one-hot vector. 

The optimization of an architecture is typically based on its validation loss in the bi-level optimization. To evaluate the performance of a sampled architecture $\mathbf{Z}=\{ \mathbf{Z}^{(i,j)} \}$, we build a target supernetwork $\bm{\hat{y}}=f(\bm{x};\bm{\omega},\mathbf{Z})$ by stacking several copies of the cells sequentially, where $\bm{x}$ is the network input and $\bm{\hat{y}}$ is the network output. The loss of the target supernetwork for classification tasks is measured by the commonly used negative log likelihood:
\begin{equation}
    \mathcal{L}\left(\bm{D};\bm{\omega},\mathbf{Z}\right) = \mathbf{E}_{(\bm{x},\bm{y})\sim \bm{D}}\left[-\log\left(f\left(\bm{x};\bm{\omega},\mathbf{Z}\right)\right)\right],
    \label{eq:loss}
\end{equation}
where $\bm{D}$ is the corresponding training, validation or testing dataset depend on the context. With the validation loss $\mathcal{L}(\bm{D}_{val};\bm{\omega},\mathbf{Z})$, the architecture weights can be jointly optimized with the target network weights by solving the  bi-level optimization problem in Eq.~\eqref{eq:bilevel_opt}.
From Eq.~\eqref{equation:onehot} and Eq.~\eqref{eq:intermediate}, the gradient of the loss $\mathcal{L}$ in Eq.~\eqref{eq:loss} w.r.t. $\alpha^{(i,j)}_k$ is:
\begin{small}
\begin{align}
    &\frac{\partial \mathcal{L}}{\partial \alpha^{(i,j)}_k} = \frac{\mathcal{L}}{\partial \mathbf{I}^{(j)}}o^{(i,j)}(\mathbf{I}^{(i)})(\delta (k-k_0)-\mathbf{Z}^{(i,j)})Z^{(i,j)}_k\frac{1}{\tau \alpha^{(i,j)}_k}.
    \label{eq:gradientofloss}
\end{align}
\end{small}

We apply the Anneal Gumbel Softmax on the operation search of the outer edges $E_{out}={\{\bigcup_{i\geq 2}(0,i),  \bigcup_{j\geq2}(1,j)\}}$ whose root node is $\mathbf{I}^{(0)}$ or $\mathbf{I}^{(1)}$ generated by the previous cells. These eight outer edges connect and transform the feature map from the previous cells to the intermediate nodes in the current cell. Since they do not contain any predecessor edge in the current cell, we treat their architecture weights as independent variables. By contrast, we introduce a new inter-layer transition learning paradigm for the inner edges $E_{in}=\bigcup_{2\leq i \leq4}(i,j)$ to derive their architecture weights explicitly, since there is a clear dependency between an inner edge and its predecessor edges.

\subsection{Inter-layer Transition Learning}
\begin{figure}
    \centering
    \includegraphics[width=1\linewidth]{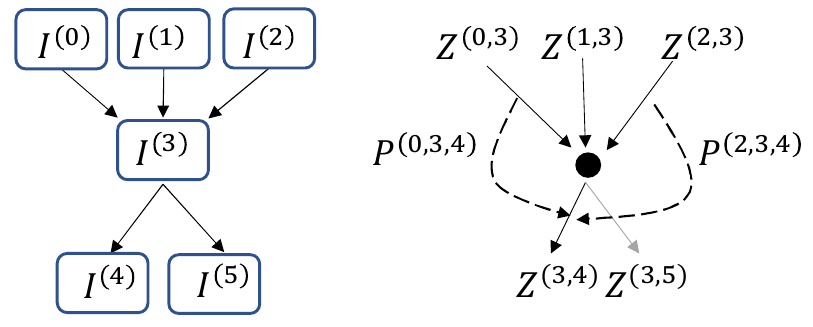} 
    \caption{Illustration of the information flow among nodes (left) and edges (right). $\mathbf{I}^{(3)}$ is generated from the mixed operations on its predecessor nodes $\mathbf{I}^{(0)}$, $\mathbf{I}^{(1)}$, and $\mathbf{I}^{(2)}$. The architecture weights of these edges are $\mathbf{Z}^{(0,3)}$, $\mathbf{Z}^{(1,3)}$, and $\mathbf{Z}^{(2,3)}$. We model the dependency between $\mathbf{Z}^{(3,4)}$ and its predecessor edges, \eg, $\mathbf{Z}^{(0,3)}$, by a learnable transition matrix $\mathbf{P}^{(0,3,4)}$.}
    \label{fig:inner_transition}
\end{figure}
DARTS and its variants consider that the architecture weights of edges are independent learnable variables. For example, as shown in Figure \ref{fig:inner_transition}, edges $(0,3),(1,3),(2,3)$ are the predecessor of the edge $(3,4)$.
DARTS and its variants optimize $\mathbf{Z}^{(3,4)}$ independently w.r.t. $\mathbf{Z}^{(0,3)}$, $\mathbf{Z}^{(1,3)}$, and $\mathbf{Z}^{(2,3)}$. However, the architecture weight $\mathbf{Z}^{(3,4)}$ is indeed influenced by the architecture weights of its predecessor edges $\mathbf{Z}^{(0,3)}$, $\mathbf{Z}^{(1,3)}$, and $\mathbf{Z}^{(2,3)}$, since the information flow is explicitly defined by the directed topological connections between these edges and influenced by their architecture weights. Thereby, the operation choice of an edge matters for the operation choices of its descendent edges.

To account for this dependency, we introduce a novel inter-layer transition learning paradigm for those inner edges. As shown in the right part of Figure~\ref{fig:inner_transition}, we define a learnable transition matrix to denote how the architecture weight of a predecessor edge (\ie, the ``state'' of operation choice) transits to the architecture weight of its descendent edge. For example, the architecture weight $\mathbf{Z}^{(i,j)}$ of edge $(i,j)$ can be deduced from the architecture weight $\mathbf{Z}^{(m,i)}$ of its predecessor edge $(m,i)$ as follows:
\begin{equation}
    \mathbf{Z}^{(i,j)}=\mathbf{P}^{(m,i,j)}\mathbf{Z}^{(m,i)}.
    \label{eq:transition}
\end{equation}
Here, $\mathbf{P}^{(m,i,j)}$ is a learnable transition matrix subjected to:
\begin{align}
    \sum_{t=0}^{K-1}p_{s,t}&=1,\forall s\in[0,K-1], \nonumber \\
    \text{s.t.}\ \ \  0 &\leq p_{s,t} \leq 1,
\end{align}
where $p_{s,t}$ is the element in $\mathbf{P}^{(m,i,j)}$, denoting the transition probability from the $s$-th operation on the edge $(m,i)$ to $t$-th operation on the edge $(i,j)$.

\begin{figure}
    \centering
    \includegraphics[width=0.8\linewidth]{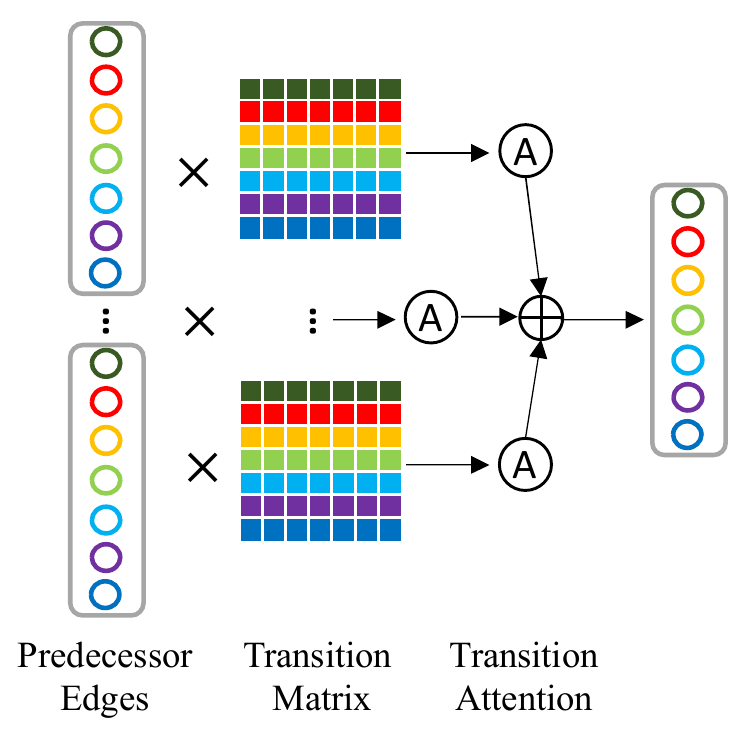}
    \caption{Illustration of the Inter-layer Transition. The architecture weight of an inner edge is deduced from those of its predecessor edges via the learnable transition matrices and attention vector.}
    \label{fig:inner_transition_sum}
\end{figure}

For an inner edge $(i,j)$, it may have multiple predecessor edges, which may have different impact on $(i,j)$. Thereby, we adopt an attention mechanism to adaptively weigh the importance of different transition matrices. As shown in Figure~\ref{fig:inner_transition_sum}, the architecture weight on an inner edge is deduced by a weighted sum of those of its predecessor edges via the learnable transition matrices and attention vectors. Mathematically, this process can be formulated as:
\begin{align}
    \mathbf{Z}^{(i,j)}=
     \sum_{i<j} \sum_{m<i} \beta^{(m,i,j)} \mathbf{P}^{(m,i,j)}\mathbf{Z}^{(m,i)}.
    \label{eq:intertransition_attention}
\end{align}
Here, $\beta^{(m,i,j)}$ denotes the attention score of the transition matrix $\mathbf{P}^{(m,i,j)}$, which is subjected to:
\begin{align}
     \sum_m \beta^{(m,i,j)} &= 1, \nonumber \\
     \text{s.t.} \ \ \ \  0 \leq \beta^{(m,i,j)} &\leq 1.
     \label{equation:beta}
\end{align}
Note that both the transition matrix $\mathbf{P}^{(m,i,j)}$ and attention score $\beta^{(m,i,j)}$ are learnable variables. In addition, since $\mathbf{Z}^{(i,j)}$ in Eq.~\eqref{eq:intertransition_attention} is differentiable w.r.t. $\mathbf{P}^{(m,i,j)}$, $\beta^{(m,i,j)}$, and $\mathbf{Z}^{(m,i)}$, thereby $\mathbf{P}^{(m,i,j)}$ and $\beta^{(m,i,j)}$ can be optimized together with the architecture weights of outer edges by minimizing the validation loss in the bi-level optimization framework. In this way, the dependency between each inner edge and its predecessor edges can be learned dynamically.

\subsection{Transition-induced Iterative Edge Pruning}

\begin{algorithm}[t]
  \caption{ Transition-induced Iterative Edge Pruning.}  
  \label{alg:iterative_pruning}  
  \begin{algorithmic}[1]
   
   \STATE \textbf{Input:} architecture weight $\mathbf{Z}^{(m,i)}$ of outer edges, transition matrix $\mathbf{P}^{(m,i,j)}$, attention $\beta^{(m,i,j)}$
   \STATE \textbf{Output:} the retained edges and operations after pruning $\bigcup_{2\leq j\leq5}(M_j,Z_j)$
   \STATE \textbf{Topology Initialization:}
    \FOR{edge $(i,j) \in E_{in}$ }
    \STATE $\mathbf{Z}^{(i,j)}=
     \sum_{i<j} \sum_{m<i} \beta^{(m,i,j)} \mathbf{P}^{(m,i,j)}\mathbf{Z}^{(m,i)}$
    \ENDFOR
    
    \STATE \textbf{Iterative Edge Pruning:}
    \STATE Initialize $M_2=\{(0,2),(1,2)\}$,  $Z_2=\{Z^{(0,2)},Z^{(1,2)}\}$
    \FOR{node $j \in$ intermediate nodes $\{3,4,5\}$}
        \FOR{$i=2,i<j,i++$}
            \IF{$(i,j) \in E_{in} $ }
               \FOR{$m=0,m<i,m++$}
                  \IF{$(m,i)\in M_i$}
                   \STATE  $\mathbf{Z}^{(m,i)} \leftarrow \text{one-hot}(\text{argmax}_k \mathbf{Z}^{(m,i)}_k))$
                  \ENDIF
               \ENDFOR
            \STATE $\mathbf{Z}^{(i,j)}=\sum_{m,i} \beta^{(m,i,j)} \mathbf{P}^{(m,i,j)}\mathbf{Z}^{(m,i)}$
        \ENDIF
        \ENDFOR
        \STATE $\{(i_0,j),(i_1,j)\} \leftarrow$ edges with top-2 $\text{max}(\mathbf{Z}^{(i,j)})$
        \STATE $M_j=\{(i_0,j),(i_1,j)\}$, $Z_j=\{Z^{(i_0,j)},Z^{(i_1,j)}\}$
    \ENDFOR
    
  \end{algorithmic}  
\end{algorithm}

After the search phase, we will get an over-parametered supernet. To generate the final architecture, the redundant operation candidates (\ie, edges) should be eliminated, aside from the crucial ones in the supernet. The importance of the edge $(i,j)$ in the searched cell is indicated by the largest architecture operation weight of $\mathbf{Z}^{(i,j)}$. The commonly used edge pruning strategy in DARTS and its variants is to retain two edges with the largest edge importance between each pair of nodes. However, this heuristic pruning strategy prunes all weak edges at once without considering the dependency between them and their descendent edges induced by the directed topological connections. To address this issue, we propose a new method named transition-induced iterative edge pruning (TIEP).

At the initialization stage for edge pruning, we can calculate the architecture weights of all outer edges $\mathbf{Z}^{(i,j)}$ according to Eq.~\eqref{eq:intertransition_attention}.
Then, we prune the redundant edge iteratively and retain two edges for each intermediate node in a cell. As shown in Figure \ref{fig:opening}, the node $\mathbf{I}^{(2)}$ is the first intermediate node and only two edges ($(0,2)$ and $(1,2)$) point to it. Thereby, these two edges will be retained. Accordingly, we select the operations on them and update the architecture weights by selecting the maximum operation probability:
\begin{align}\nonumber
    \mathbf{Z}^{(i,j)} &\leftarrow \text{one-hot}\left(k^{*}=\text{argmax}_k \mathbf{Z}^{(i,j)}_k\right),\\
    o^{(i,j)} &\leftarrow o_{k^{*}}, i<j, j=2.
    \label{eq:one_hot_pruning}
\end{align}
For node $\mathbf{I}^{(3)}$, there are three edges pointing to it, including $(0,3),(1,3)$, and $(2,3)$. The edges $(0,3)$ and $(1,3)$ are outer edges while the edge $(2,3)$ is an inner edge, whose architecture weight is calculated according to Eq.~\eqref{eq:intertransition_attention}, \ie,
\begin{equation}
    \mathbf{Z}^{(2,3)} =\beta^{(0,2,3)}\mathbf{P}^{(0,2,3)}\mathbf{Z}^{(0,2)}+\beta^{(1,2,3)}\mathbf{P}^{(1,2,3)}\mathbf{Z}^{(1,2)}.
    \label{equation:sample_inner}
\end{equation}
Note that $\mathbf{Z}^{(0,2)}$ and $\mathbf{Z}^{(1,2)}$ have been updated as the one-hot vectors according to Eq.\eqref{eq:one_hot_pruning} in the previous pruning phase. The attention score $\beta^{(0,2,3)}$ and $\beta^{(1,2,3)}$ and the transition matrices $\mathbf{P}^{(0,2,3)}$ and $\mathbf{P}^{(1,2,3)}$ have been learned during the architecture search phase. Then we prune the edge $(r,3)$ subject to $r={\text{argmin}}_{i}({\text{max}}({Z}^{(i,3)}))$. Since the pruned edge may be the predecessor edge of other inner edges, their architecture weights should be updated to account for the absence of this predecessor edge according to Eq.~\eqref{eq:intertransition_attention}. For example, the transition matrix $\mathbf{P}^{(r,3,j)}$ and attention score $\beta^{(r,3,j)}$ will become zero due to the removal of edge $(r,3)$, if edge $(3,j)$ is a descendent inner edge of the edge $(r,3)$. Then, we reweight the attention score from edge $(m,3)$ to edge $(3,j)$ by softmax and calculate the architecture weight of the edge $(3,j)$ as follows: 
\begin{align}
 \mathbf{Z}^{(3,j)} &=
     \sum_{3<j} \sum_{m\neq r} \beta^{(m,3,j)} \mathbf{P}^{(m,3,j)}\mathbf{Z}^{(m,3)}, \nonumber \\
     &\text{s.t.} \ \sum_{m\neq r} \beta^{(m,3,j)} = 1.
\end{align}
Since there are only two edges left for the node $\mathbf{I}^{(3)}$ after pruning the edge $(r,3)$, they will be retained and their architecture weights should be updated to the one-hot vectors similar to Eq.~\eqref{eq:one_hot_pruning}. In this way, we can iteratively prune those redundant and unimportant edges while updating the architecture weight $\mathbf{Z}$ at the same time. The complete process of the transition-induced iterative edge pruning method is illustrated in Algorithm~\ref{alg:iterative_pruning}.

\section{Experiments}
We evaluated the performance of architectures searched by ITNAS on the CIFAR-10 \cite{krizhevsky2009learning} and we further evaluated their generalization ability on other popular image classification benchmarks, including the widely used ImageNet \cite{deng2009imagenet}, CIFAR-100 \cite{krizhevsky2009learning}, CINIC-10 \cite{darlow2018cinic}, and SVHN \cite{netzer2011reading}. 

\subsection{Results on CIFAR-10}
\subsubsection{Architecture Search and Evaluation}

Following the common protocol in previous work, we used the CIFAR-10 dataset for architecture search and evaluation. CIFAR-10 contains $50,000$ training images together with $10,000$ testing images from $10$ classes. During the search phase, we shuffled the training set and divide it into two parts of equal size for training network weights and architecture weights, respectively.

We set two types of cells in the search space, \ie, normal cell and reduction cell. The operations in the normal cells preserve the spatial dimensions, while the operations on the edges connected to the input nodes in the reduction cells have a stride of two, and the spatial solution of the output features is halved. We set the number of candidate operations to $K=7$ to include four convolutional operations (\ie, $3\times3$ separable convolution, $5\times5$ separable convolution, $3\times3$ dilated separable convolution, and $5\times5$ dilated separable convolution), two pooling operations (\ie, $3\times3$ average pooling and $3\times3$ max pooling), and a special $identity$ operation denoting the skip connection. The target supernet to be searched is constructed by stacking six normal cells and two reduction cells. The $3rd$ and $6th$ cells are reduction cells, while others are normal cells. The initial number of feature channels is $16$ and doubled after each reduction cell.

We trained the supernet on CIFAR-10 for $50$ epochs. To optimize the network weight $\bm{\omega}$, we used the SGD optimizer. The initial learning rate is $0.025$ , which is gradually decreased to $1e-3$ using cosine annealing. The Softmax temperature $\tau$ in Gumbel-Softmax is initialized as $5$ and linearly reduced to $0.5$. The architecture weights $\bm{\alpha}$ of outer edges, the transition matrix $\mathbf{P}$, and attention vector $\bm{\beta}$ are jointly optimized using the Adam optimizer with a learning rate of $3e-4$ and a weight decay of $1e-3$.

For a fair comparison, every architecture shares the same hyper-parameters adopted from DARTS for training. During the evaluation, $20$ searched cells are stacked to form a larger network, which is then trained from scratch for $600$ epochs with batch size of $96$. The learning rate is initialized as $0.025$ and decreased to $0$ using cosine annealing. The probability of path drop is $0.2$, and the weight of the auxiliary classifier is $0.4$. We used the common pre-possessing and data augmentation techniques, \ie, randomly cropping, horizontally flipping, normalization, and cutout \cite{devries2017improved}.

\begin{table}[htbp]
  \centering
    \begin{tabular}{lcccc}
    \toprule
    \textbf{Architecture                 } & \multicolumn{1}{c}{\textbf{Error  (\%)}} & \multicolumn{1}{c}{\textbf{Params (M)}} & \multicolumn{1}{c}{\textbf{Cost}}\\
    \midrule
    DenseNet-BC \cite{huang2017densely}& 3.46  & 25.6  & -   \\
    \midrule
    NASNet-A \cite{zoph2018learning} & 2.65  & 3.3   & 1800  \\
    AmoebaNet-A \cite{real2019regularized}& 3.34  & 3.2   & 3150 \\
    AmoebaNet-B \cite{real2019regularized}& 2.55  & 2.8   & 3150 \\
    Hier-Evolution \cite{liu2018hierarchical} & 3.75  & 15.7  & 300  \\
    PNAS \cite{liu2018progressive} & 3.41  & 3.2   & 225   \\
    ENAS \cite{pham2018efficient} & 2.89  & 4.6   & 0.50  \\
    NAONet-WS \cite{luo2018neural} & 3.53  & 3.1   & 0.40   \\
    \midrule
    DARTS (1st) \cite{liu2018darts} & 3.00  & 3.3   & 0.4   \\
    DARTS (2nd) \cite{liu2018darts} & 2.76  & 3.3   & 1.0     \\
    SNAS \cite{xie2018snas} & 2.98  & 2.9   & 1.5   \\
    GDAS \cite{dong2019searching} & 2.93  & 3.4   & 0.21   \\
    BayesNAS \cite{zhou2019bayesnas} & 2.81  & 3.4   & 0.20   \\
    PVLL-NAS \cite{li2020neural}& 2.70  & 3.3   & 0.20   \\
    MiLeNAS \cite{he2020milenas} & 2.51 & 3.9 & 0.30 \\
    SGAS (Cri.1) \cite{li2020sgas} & 2.66  & 3.7   & 0.25  \\
    SGAS (Cri.2) \cite{li2020sgas}&2.52  & 4.1   & 0.25   \\
    \textbf{ITNAS} & \textbf{2.45}  & 4.0   & 0.30   \\
    \bottomrule
    \end{tabular}%
    \caption{Comparison between ITNAS and state-of-the-art methods on CIFAR-10. The searching cost is measured as GPU days.}
  \label{tab:cifar10}%
\end{table}%
We compared the network searched by our ITNAS with both hand-crafted ones and those searched by state-of-the-art (SOTA) NAS methods in Table \ref{tab:cifar10}. ITNAS outperforms all previous methods with comparable amount of parameters. 
It discovers an architecture with $4.0$M parameters in $0.3$ GPU day, which achieves $2.45\%$ classification error on CIFAR-10. The number of architecture parameters optimized in ITNAS is slightly more than DARTS (\ie, only increasing $1376$ parameters) by introducing the transition matrices and attention vectors. The dependency of architecture weights between inner edges and their predecessor edges are modeled effectively by these parameters. Intuitively, the proposed inter-layer transition learning paradigm can regularize the search space of inner edges and explicitly influence the optimization of the operation choices on outer edges via error back-propagation according to Eq.~\eqref{eq:intertransition_attention}. Thereby, better architectures can be discovered.
\begin{table}[htbp]
  \centering
    \begin{tabular}{cccc}
    \toprule
    \textbf{Gumbel}& \multicolumn{1}{c}{\textbf{Transition} } & \multicolumn{1}{c}{\textbf{TIEP}} & \multicolumn{1}{c}{\textbf{Error (\%)}}\\
    \midrule

    \CheckmarkBold & \XSolidBrush  & \XSolidBrush   & 2.72  \\
    \CheckmarkBold & \CheckmarkBold  & \XSolidBrush   & 2.53  \\
    \CheckmarkBold & \CheckmarkBold  & \CheckmarkBold   &  2.45 \\
    \bottomrule
    \end{tabular}%
    \caption{Ablation study of ITNAS on CIFAR-10, \ie, Gumbel: Annealing Gumbel Softmax, Transition: Inter-layer Transition Learning, TIEP: Transition-induced Iterative Edge Pruning.}
  \label{tab:ablation}%
\end{table}%
\subsubsection{Ablation Study}.
When we remove all these three modules, \ie, Annealing Gumbel Softmax, Inter-layer Transition Learning, and TIEP, our ITNAS will degenerate to DARTS, which searches an architecture with $2.76\%$ classification error. As shown in Table~\ref{tab:ablation}, each of these modules contributes to the gains of ITNAS over DARTS, \ie, the classification error is reduced to $2.72\%$, $2.53\%$, and $2.45\%$, after incorporating them one-by-one. The inter-layer transition learning brings more gains than the other two modules. Nevertheless, the proposed pruning method TIEP is also effective since it reduces the classification error further considering that the score has been saturated on CIFAR-10. It is noteworthy that this new pruning method is also induced by the inter-layer transition learning paradigm. Thereby, these results confirm the value of modeling the inter-layer dependency for architecture search.

\subsection{Result on ImageNet}
\begin{figure}
    \centering
    \includegraphics[width=0.8\linewidth]{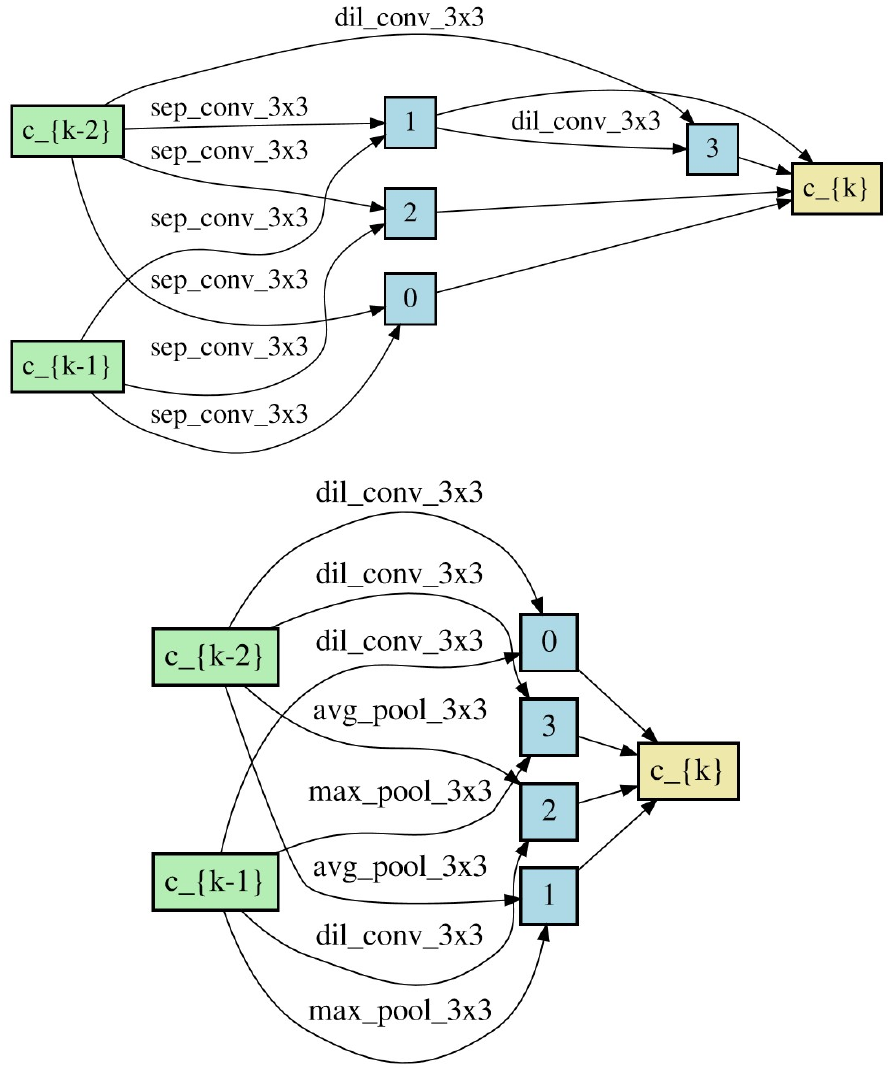} 
    \caption{Illustration of the searched normal (the upper one) and reduction (the lower one) cells.}
    \label{fig:searched_cell}
\end{figure}
\begin{table*}[htbp]
  \centering
  
    \begin{tabular}{lcccccc}
    \toprule
    \textbf{Architecture                 } & \multicolumn{1}{c}{\textbf{Top-1  (\%)}} & \multicolumn{1}{c}{\textbf{Top-5  (\%)}} & \multicolumn{1}{c}{\textbf{Params (M)}} & \multicolumn{1}{c}{\textbf{FLOPs  (M)}} & \multicolumn{1}{c}{\textbf{Cost}} & \multicolumn{1}{c}{\textbf{Method}} \\
    \midrule
    Inception-v1 \cite{szegedy2015going}& 30.2  & 10.1  & 6.6   & 1448  & -     & manual \\
    MobileNet \cite{howard2017mobilenets}& 29.4  & 10.5  & 4.2   & 569   & -     & manual \\
    ShuffleNet 2x (v1) \cite{zhang2018shufflenet}& 26.4  & 10.2  & 5     & 524   & -     & manual \\
    ShuffleNet 2x (v2) \cite{ma2018shufflenet}& 25.1  & -     & 5     & 591   & -     & manual \\
    \midrule
    NASNet-A \cite{zoph2018learning}& 26.0    & 8.4   & 5.3   & 564   & 1800  & RL \\
    NASNet-B \cite{zoph2018learning}& 27.2  & 8.7   & 5.3   & 488   & 1800  & RL \\
    NASNet-C \cite{zoph2018learning}& 27.5  & 8.9   & 4.9   & 558   & 1800  & RL \\
    AmoebaNet-A \cite{real2019regularized}& 25.5  & 8     & 5.1   & 555   & 3150  & EA \\
    AmoebaNet-B \cite{real2019regularized}& 26.0    & 8.5   & 5.3   & 555   & 3150  & EA \\
    AmoebaNet-C \cite{real2019regularized}& 24.3  & 7.6   & 6.4   & 570   & 3150  & EA \\
    PNAS \cite{liu2018progressive}  & 25.8  & 8.1   & 5.1   & 588   & 225   & SMBO \\
    \midrule
    DARTS(2nd) \cite{liu2018darts} & 26.7  & 8.7   & 4.7   & 574   & 4     & gradient \\
    PDARTS \cite{chen2019progressive}& 24.4  & 7.4   & 4.9   & 557   & 0.3   & gradient \\
    SNAS \cite{xie2018snas} & 27.3  & 9.2   & 4.3   & 522   & 1.5   & gradient \\
    ProxylessNAS \cite{chen2019progressive}& 24.9  & 7.5   & 7.1   & 465   & 8.3   & gradient \\
    GDAS \cite{dong2019searching} & 26.0  & 8.5   & 5.3   & 581   & 0.21  & gradient \\
    PVLL-NAS \cite{li2020neural} & 25.60  & 8.10  & 5.0     & 556   & 0.20   & gradient \\
    CARS-I \cite{yang2020cars}& 24.80  & 7.50   & 5.1   & 591   & 0.40   & evolution \\
    MiLeNAS \cite{he2020milenas} & 24.70 &7.60 & 5.3 & 584 & 0.30 &gradient \\
    SGAS (Cri.1) \cite{li2020sgas}  & 24.41 & 7.29  & 5.3   & 579   & 0.25  & gradient \\
    SGAS (Cri.2) \cite{li2020sgas} & 24.44 & 7.41  & 5.7   & 621   & 0.25  & gradient \\
    \textbf{ITNAS} &  \textbf{24.04}  & \textbf{7.29}  & 5.5  &   627    &    0.30   & gradient \\
    \bottomrule
    \end{tabular}%
    \caption{Comparison between ITNAS and state-of-the-art methods on ImageNet.}
  \label{tab:imagenet}%
\end{table*}%
The normal and reduction cells obtained by ITNAS are illustrated in Figure \ref{fig:searched_cell}. We evaluated their transferability on the large-scale ImageNet dataset \cite{deng2009imagenet}. ImageNet contains 1.3 million training images and $50,000$ testing images from $1,000$ categories. We followed the common single-crop evaluation protocol, where the input size is $224\times224$. Training hyper-parameters are adopted from PC-DARTS \cite{xu2019pc}. The final network on ImageNet has $14$ layers, and the initial number of feature channels is $48$. It is trained for $250$ epochs with a batch size of $1024$. The SGD optimizer is used with an initial learning rate of $0.5$, which is decreased to 0 linearly. The momentum and weight decay are $0.9$ and $3e-5$. Learning rate warm-up is used in the first $5$ epochs.
 
In Table \ref{tab:imagenet}, we compared ITNAS with SOTA NAS methods and manually designed networks on ImageNet. It reveals that the networks searched by NAS are superior to those manually designed. In addition, the differential NAS methods significantly reduce the search cost compared with RL-based and EA-based ones. Our model achieves a top-1 error of $24.04\%$, outperforming all the others. It has comparable amount of parameters and computation complexity as other SOTA NAS methods, \ie, $5.5$M parameters and 627M FLOPs. It is searched by ITNAS in $0.3$ GPU day, retaining the efficiency of gradient-based NAS methods. Note that SGAS \cite{li2020sgas} chooses and prunes candidate operations in a greedy fashion and obtains good results. Nevertheless, it also ignores the dependency between edges. By contrast, our ITNAS benefits from modeling such dependency by learning inter-layer transition and outperforms SGAS.

\begin{table}[htbp]
  \centering
    \begin{tabular}{lcccc}
    \toprule
    \textbf{Architecture                 } & \multicolumn{1}{c}{CINIC-10} & \multicolumn{1}{c}{CIFAR-100} & \multicolumn{1}{c}{SVHN}\\
    \midrule
    Known SOTA & 8.6 \cite{darlow2018cinic}  & 8.7 \cite{huang2019gpipe} & 1.02 \cite{cubuk2019autoaugment}  \\
    \midrule
    NASNet \cite{zoph2018learning}& 6.93  & 15.8  & 1.96  \\
    AmoebaNet-A \cite{real2019regularized}& 7.18  & 15.9  & 1.93 \\
    PNAS \cite{liu2018progressive}& 7.03  & 15.9   & 1.83   \\
    SNAS \cite{xie2018snas}& 7.13  & 16.5   & 1.98  \\
    DARTS(1st) \cite{liu2018darts}& 7.05  &15.8   &1.94   \\
    DARTS(2nd) \cite{liu2018darts}& 6.88  & 15.7   & 1.85     \\
    ASAP \cite{noy2020asap} & 6.83  & 15.6  & 1.81  \\
    \textbf{ITNAS} & \textbf{6.72}  &  \textbf{15.5}  & \textbf{1.78} \\
    \bottomrule
    \end{tabular}%
    \caption{Transferability of the searched cells by ITNAS and SOTA NAS methods on CNIC-10 \cite{darlow2018cinic}, CIFAR-100 \cite{krizhevsky2009learning} and SVHN \cite{netzer2011reading}.}
  \label{tab:other_datasets}%
\end{table}%
\subsection{Results on other three benchmark datasets}
We further evaluated the transferability of the searched cells by ITNAS on other three benchmarks, \ie, CNIC-10 \cite{darlow2018cinic}, CIFAR-100 \cite{krizhevsky2009learning} and SVHN \cite{netzer2011reading}. CINIC-10 is an extension of CIFAR-10 via the addition of downsampled ImageNet images. It contains $270,000$ images of $10$ classes. CIFAR-100 contains $60,000$ images from $100$ classes, including $50,000$ images for training and $10,000$ images for test. The image size is $32\times32$. SVHN contains $600,000$ $32\times32$ real-world images of $10$ classes of digits and numbers from natural scenes. For a fair comparison, we adopted the same network configuration and hyper-parameters as ASAP \cite{noy2020asap}, except from the searched cells. The results of ITNAS and SOTA methods are summarized in Table~\ref{tab:other_datasets}, showing that the cells searched by ITNAS can generalize well on other classification tasks and outperforms other cells searched by SOTA methods. In addition, ITNAS sets a new SOTA on CINIC-10, \ie, $6.72\%$ classification error.

\section{Conclusion}
In this paper, we demonstrate that modeling the dependency between edges in the DAG matters for NAS. We propose a novel inter-layer transition NAS method (ITNAS) to explicitly model such dependency in an attentive probability transition manner. In addition, a new iterative edge pruning method is derived naturally from the inter-layer transition paradigm, which can iteratively prune redundant searched edges and dynamically updating architecture weights. ITNAS outperforms state-of-the-art NAS methods on five benchmarks. This study makes the first attempt to model inter-layer dependency and shows promising results, though more efforts are expected to advance NAS in this direction. For example, other forms of dependency formulation can be attempted, \eg, learning a nonlinear mapping by a neural module. Besides, pruning unimportant edges guided by the dependency information during search is also worth studying to improve the searching efficiency.
 

{\small
\bibliographystyle{ieee_fullname}
\bibliography{cvpr}

\begin{thebibliography}{10}\itemsep=-1pt

\bibitem{baker2016designing}
Bowen Baker, Otkrist Gupta, Nikhil Naik, and Ramesh Raskar.
\newblock Designing neural network architectures using reinforcement learning.
\newblock {\em International Conference on Learning Representations, (ICLR)},
  2017.

\bibitem{chen2019progressive}
Xin Chen, Lingxi Xie, Jun Wu, and Qi Tian.
\newblock Progressive differentiable architecture search: Bridging the depth
  gap between search and evaluation.
\newblock In {\em IEEE International Conference on Computer Vision, (ICCV)},
  2019.

\bibitem{cubuk2019autoaugment}
Ekin~D Cubuk, Barret Zoph, Dandelion Mane, Vijay Vasudevan, and Quoc~V Le.
\newblock Autoaugment: Learning augmentation strategies from data.
\newblock In {\em IEEE Conference on Computer Vision and Pattern Recognition,
  (CVPR)}, 2019.

\bibitem{darlow2018cinic}
Luke~N Darlow, Elliot~J Crowley, Antreas Antoniou, and Amos~J Storkey.
\newblock Cinic-10 is not imagenet or cifar-10.
\newblock {\em arXiv preprint arXiv:1810.03505}, 2018.

\bibitem{deng2009imagenet}
Jia Deng, Wei Dong, Richard Socher, Li-Jia Li, Kai Li, and Li Fei-Fei.
\newblock Imagenet: A large-scale hierarchical image database.
\newblock In {\em IEEE Conference on Computer Vision and Pattern Recognition,
  (CVPR)}, 2009.

\bibitem{devries2017improved}
Terrance DeVries and Graham~W Taylor.
\newblock Improved regularization of convolutional neural networks with cutout.
\newblock In {\em arXiv preprint arXiv:1708.04552}, 2017.

\bibitem{dong2019searching}
Xuanyi Dong and Yi Yang.
\newblock Searching for a robust neural architecture in four gpu hours.
\newblock In {\em IEEE Conference on Computer Vision and Pattern Recognition,
  (CVPR)}, 2019.

\bibitem{elsken2019neural}
Thomas Elsken, Jan~Hendrik Metzen, and Frank Hutter.
\newblock Neural architecture search: A survey.
\newblock {\em Journal of Machine Learning Research, (JMLR)}, 20:1--21, 2019.

\bibitem{fang2020densely}
Jiemin Fang, Yuzhu Sun, Qian Zhang, Yuan Li, Wenyu Liu, and Xinggang Wang.
\newblock Densely connected search space for more flexible neural architecture
  search.
\newblock In {\em IEEE Conference on Computer Vision and Pattern Recognition,
  (CVPR)}, 2020.

\bibitem{guo2020hit}
Jianyuan Guo, Kai Han, Yunhe Wang, Chao Zhang, Zhaohui Yang, Han Wu, Xinghao
  Chen, and Chang Xu.
\newblock Hit-detector: Hierarchical trinity architecture search for object
  detection.
\newblock In {\em IEEE Conference on Computer Vision and Pattern Recognition,
  (CVPR)}, 2020.

\bibitem{he2020milenas}
Chaoyang He, Haishan Ye, Li Shen, and Tong Zhang.
\newblock Milenas: Efficient neural architecture search via mixed-level
  reformulation.
\newblock In {\em IEEE Conference on Computer Vision and Pattern Recognition,
  (CVPR)}, 2020.

\bibitem{he2016deep}
Kaiming He, Xiangyu Zhang, Shaoqing Ren, and Jian Sun.
\newblock Deep residual learning for image recognition.
\newblock In {\em IEEE Conference on Computer Vision and Pattern Recognition,
  (CVPR)}, 2016.

\bibitem{howard2017mobilenets}
Andrew~G Howard, Menglong Zhu, Bo Chen, Dmitry Kalenichenko, Weijun Wang,
  Tobias Weyand, Marco Andreetto, and Hartwig Adam.
\newblock Mobilenets: Efficient convolutional neural networks for mobile vision
  applications.
\newblock {\em arXiv preprint arXiv:1704.04861}, 2017.

\bibitem{hu2018squeeze}
Jie Hu, Li Shen, and Gang Sun.
\newblock Squeeze-and-excitation networks.
\newblock In {\em IEEE Conference on Computer Vision and Pattern Recognition,
  (CVPR)}, 2018.

\bibitem{huang2017densely}
Gao Huang, Zhuang Liu, Laurens Van Der~Maaten, and Kilian~Q Weinberger.
\newblock Densely connected convolutional networks.
\newblock In {\em IEEE Conference on Computer Vision and Pattern Recognition,
  (CVPR)}, 2017.

\bibitem{huang2019gpipe}
Yanping Huang, Youlong Cheng, Ankur Bapna, Orhan Firat, Dehao Chen, Mia Chen,
  HyoukJoong Lee, Jiquan Ngiam, Quoc~V Le, Yonghui Wu, et~al.
\newblock Gpipe: Efficient training of giant neural networks using pipeline
  parallelism.
\newblock In {\em Neural Information Processing Systems, (NIPS)}, 2019.

\bibitem{krizhevsky2009learning}
Alex Krizhevsky, Geoffrey Hinton, et~al.
\newblock Learning multiple layers of features from tiny images.
\newblock {\em Citeseer, Tech. Rep}, 2009.

\bibitem{li2020sgas}
Guohao Li, Guocheng Qian, Itzel~C Delgadillo, Matthias Muller, Ali Thabet, and
  Bernard Ghanem.
\newblock Sgas: Sequential greedy architecture search.
\newblock In {\em IEEE Conference on Computer Vision and Pattern Recognition,
  (CVPR)}, 2020.

\bibitem{li2020neural}
Yanxi Li, Minjing Dong, Yunhe Wang, and Chang Xu.
\newblock Neural architecture search in a proxy validation loss landscape.
\newblock In {\em International Conference on Machine Learning, (ICML)}, 2020.

\bibitem{liu2019auto}
Chenxi Liu, Liang-Chieh Chen, Florian Schroff, Hartwig Adam, Wei Hua, Alan~L
  Yuille, and Li Fei-Fei.
\newblock Auto-deeplab: Hierarchical neural architecture search for semantic
  image segmentation.
\newblock In {\em IEEE Conference on Computer Vision and Pattern Recognition,
  (CVPR)}, 2019.

\bibitem{liu2018progressive}
Chenxi Liu, Barret Zoph, Maxim Neumann, Jonathon Shlens, Wei Hua, Li-Jia Li, Li
  Fei-Fei, Alan Yuille, Jonathan Huang, and Kevin Murphy.
\newblock Progressive neural architecture search.
\newblock In {\em European Conference on Computer Vision, (ECCV)}, 2018.

\bibitem{liu2018hierarchical}
Hanxiao Liu, Karen Simonyan, Oriol Vinyals, Chrisantha Fernando, and Koray
  Kavukcuoglu.
\newblock Hierarchical representations for efficient architecture search.
\newblock In {\em International Conference on Learning Representations,
  (ICLR)}, 2018.

\bibitem{liu2018darts}
Hanxiao Liu, Karen Simonyan, and Yiming Yang.
\newblock Darts: Differentiable architecture search.
\newblock In {\em International Conference on Learning Representations,
  (ICLR)}, 2018.

\bibitem{luo2018neural}
Renqian Luo, Fei Tian, Tao Qin, Enhong Chen, and Tie-Yan Liu.
\newblock Neural architecture optimization.
\newblock In {\em Neural Information Processing Systems, (NIPS)}, 2018.

\bibitem{ma2018shufflenet}
Ningning Ma, Xiangyu Zhang, Hai-Tao Zheng, and Jian Sun.
\newblock Shufflenet v2: Practical guidelines for efficient cnn architecture
  design.
\newblock In {\em European conference on computer vision, (ECCV)}, 2018.

\bibitem{maddison2016concrete}
Chris~J Maddison, Andriy Mnih, and Yee~Whye Teh.
\newblock The concrete distribution: A continuous relaxation of discrete random
  variables.
\newblock {\em International Conference on Learning Representations, (ICLR)},
  2017.

\bibitem{nekrasov2019fast}
Vladimir Nekrasov, Hao Chen, Chunhua Shen, and Ian Reid.
\newblock Fast neural architecture search of compact semantic segmentation
  models via auxiliary cells.
\newblock In {\em IEEE Conference on Computer Vision and Pattern Recognition,
  (CVPR)}, 2019.

\bibitem{netzer2011reading}
Yuval Netzer, Tao Wang, Adam Coates, Alessandro Bissacco, Bo Wu, and Andrew~Y
  Ng.
\newblock Reading digits in natural images with unsupervised feature learning.
\newblock In {\em Neural Information Processing Systems Workshop, (NIPS
  Workshop)}, 2011.

\bibitem{noy2020asap}
Asaf Noy, Niv Nayman, Tal Ridnik, Nadav Zamir, Sivan Doveh, Itamar Friedman,
  Raja Giryes, and Lihi Zelnik.
\newblock Asap: Architecture search, anneal and prune.
\newblock In {\em International Conference on Artificial Intelligence and
  Statistics, (AISTATS)}, 2020.

\bibitem{pham2018efficient}
Hieu Pham, Melody Guan, Barret Zoph, Quoc Le, and Jeff Dean.
\newblock Efficient neural architecture search via parameters sharing.
\newblock In {\em International Conference on Machine Learning, (ICML)}, 2018.

\bibitem{real2019regularized}
Esteban Real, Alok Aggarwal, Yanping Huang, and Quoc~V Le.
\newblock Regularized evolution for image classifier architecture search.
\newblock In {\em AAAI Conference on Artificial Intelligence, (AAAI)}, 2019.

\bibitem{real2017large}
Esteban Real, Sherry Moore, Andrew Selle, Saurabh Saxena, Yutaka~Leon Suematsu,
  Jie Tan, Quoc~V Le, and Alexey Kurakin.
\newblock Large-scale evolution of image classifiers.
\newblock In {\em International Conference on Machine Learning, (ICML)}, 2017.

\bibitem{szegedy2015going}
Christian Szegedy, Wei Liu, Yangqing Jia, Pierre Sermanet, Scott Reed, Dragomir
  Anguelov, Dumitru Erhan, Vincent Vanhoucke, and Andrew Rabinovich.
\newblock Going deeper with convolutions.
\newblock In {\em IEEE Conference on Computer Vision and Pattern
  Recognition,(CVPR)}, 2015.

\bibitem{tan2020efficientdet}
Mingxing Tan, Ruoming Pang, and Quoc~V Le.
\newblock Efficientdet: Scalable and efficient object detection.
\newblock In {\em IEEE Conference on Computer Vision and Pattern Recognition,
  (CVPR)}, 2020.

\bibitem{xie2017aggregated}
Saining Xie, Ross Girshick, Piotr Doll{\'a}r, Zhuowen Tu, and Kaiming He.
\newblock Aggregated residual transformations for deep neural networks.
\newblock In {\em IEEE Conference on Computer Vision and Pattern Recognition,
  (CVPR)}, 2017.

\bibitem{xie2018snas}
Sirui Xie, Hehui Zheng, Chunxiao Liu, and Liang Lin.
\newblock Snas: stochastic neural architecture search.
\newblock In {\em International Conference on Learning Representations,
  (ICLR)}, 2018.

\bibitem{xu2019auto}
Hang Xu, Lewei Yao, Wei Zhang, Xiaodan Liang, and Zhenguo Li.
\newblock Auto-fpn: Automatic network architecture adaptation for object
  detection beyond classification.
\newblock In {\em IEEE Conference on Computer Vision and Pattern Recognition,
  (CVPR)}, 2019.

\bibitem{xu2019pc}
Yuhui Xu, Lingxi Xie, Xiaopeng Zhang, Xin Chen, Guo-Jun Qi, Qi Tian, and
  Hongkai Xiong.
\newblock Pc-darts: Partial channel connections for memory-efficient
  architecture search.
\newblock In {\em International Conference on Learning Representations,
  (ICLR)}, 2019.

\bibitem{yang2020cars}
Zhaohui Yang, Yunhe Wang, Xinghao Chen, Boxin Shi, Chao Xu, Chunjing Xu, Qi
  Tian, and Chang Xu.
\newblock Cars: Continuous evolution for efficient neural architecture search.
\newblock In {\em IEEE Conference on Computer Vision and Pattern Recognition,
  (CVPR)}, 2020.

\bibitem{zhang2018shufflenet}
Xiangyu Zhang, Xinyu Zhou, Mengxiao Lin, and Jian Sun.
\newblock Shufflenet: An extremely efficient convolutional neural network for
  mobile devices.
\newblock In {\em IEEE Conference on Computer Vision and Pattern Recognition,
  (CVPR)}, 2018.

\bibitem{zhong2018practical}
Zhao Zhong, Junjie Yan, Wei Wu, Jing Shao, and Cheng-Lin Liu.
\newblock Practical block-wise neural network architecture generation.
\newblock In {\em IEEE Conference on Computer Vision and Pattern Recognition,
  (CVPR)}, 2018.

\bibitem{zhou2019bayesnas}
H Zhou, M Yang, J Wang, and W Pan.
\newblock Bayesnas: A bayesian approach for neural architecture search.
\newblock In {\em International Conference on Machine Learning, (ICML)}, 2019.

\bibitem{zoph2016neural}
Barret Zoph and Quoc~V Le.
\newblock Neural architecture search with reinforcement learning.
\newblock {\em International Conference on Learning Representations, (ICLR)},
  2017.

\bibitem{zoph2018learning}
Barret Zoph, Vijay Vasudevan, Jonathon Shlens, and Quoc~V Le.
\newblock Learning transferable architectures for scalable image recognition.
\newblock In {\em IEEE Conference on Computer Vision and Pattern Recognition,
  (CVPR)}, 2018.

\end{thebibliography}
}
\typeout{get arXiv to do 4 passes: Label(s) may have changed. Rerun}
\end{document}